\pdfoutput=1

\documentclass[11pt]{article}

\usepackage[preprint]{acl}

\usepackage{times}
\usepackage{latexsym}

\usepackage[T1]{fontenc}

\usepackage[utf8]{inputenc}

\usepackage{microtype}

\usepackage{inconsolata}

\usepackage{graphicx}

\usepackage{amsmath}
\usepackage{amssymb}
\usepackage{changes}
\usepackage{tabularx}
\usepackage{booktabs}  

\title{Learning to Play Like Humans: A Framework for LLM Adaptation in Interactive Fiction Games}

\author{Jinming Zhang \ Yunfei Long \\
    University of Essex \\
  \texttt{jz22273@essex.ac.uk} \ \texttt{yl20051@essex.ac.uk} \\}

\author{Jinming Zhang \\
  University of Essex, UK \\
  \texttt{jz22273@essex.ac.uk} \\\And
  Yunfei Long \\
  Queen Mary University of London, UK \\
  \texttt{qp241311@qmul.ac.uk} \\}

\begin{document}
\maketitle
\begin{abstract}

Interactive Fiction games (IF games) are where players interact through natural language commands. While recent advances in Artificial Intelligence agents have reignited interest in IF games as a domain for studying decision-making, existing approaches prioritize task-specific performance metrics over human-like comprehension of narrative context and gameplay logic. This work presents a cognitively inspired framework that guides Large Language Models (LLMs) to learn and play IF games systematically. Our proposed \textbf{L}earning to \textbf{P}lay \textbf{L}ike \textbf{H}umans (LPLH) framework integrates three key components: (1) structured map building to capture spatial and narrative relationships, (2) action learning to identify context-appropriate commands, and (3) feedback-driven experience analysis to refine decision-making over time. By aligning LLMs-based agents' behavior with narrative intent and commonsense constraints, LPLH moves beyond purely exploratory strategies to deliver more interpretable, human-like performance. Crucially, this approach draws on cognitive science principles to more closely simulate how human players read, interpret, and respond within narrative worlds. As a result, LPLH reframes the IF games challenge as a learning problem for LLMs-based agents, offering a new path toward robust, context-aware gameplay in complex text-based environments.
\end{abstract}

\section{Introduction}

Interactive Fiction games (IF games), originating in the 1970s \cite{spring2015gaming, aarseth1995cybertext}, demanded abstract reasoning, implicit world inference, and narrative reconstruction from textual cues alone. Unlike visual or auditory games, IF games rely solely on language and imagination. Successful play involves iterative exploration, learning, and adaptation, guided by intuition, pattern recognition, and experience-driven generalization \cite{zander2016intuition}. Consequently, IF games offer a rich testbed for agent problem-solving, shedding light on core mechanisms of exploration and learning.


\begin{figure}[t]
  \includegraphics[width=\columnwidth]{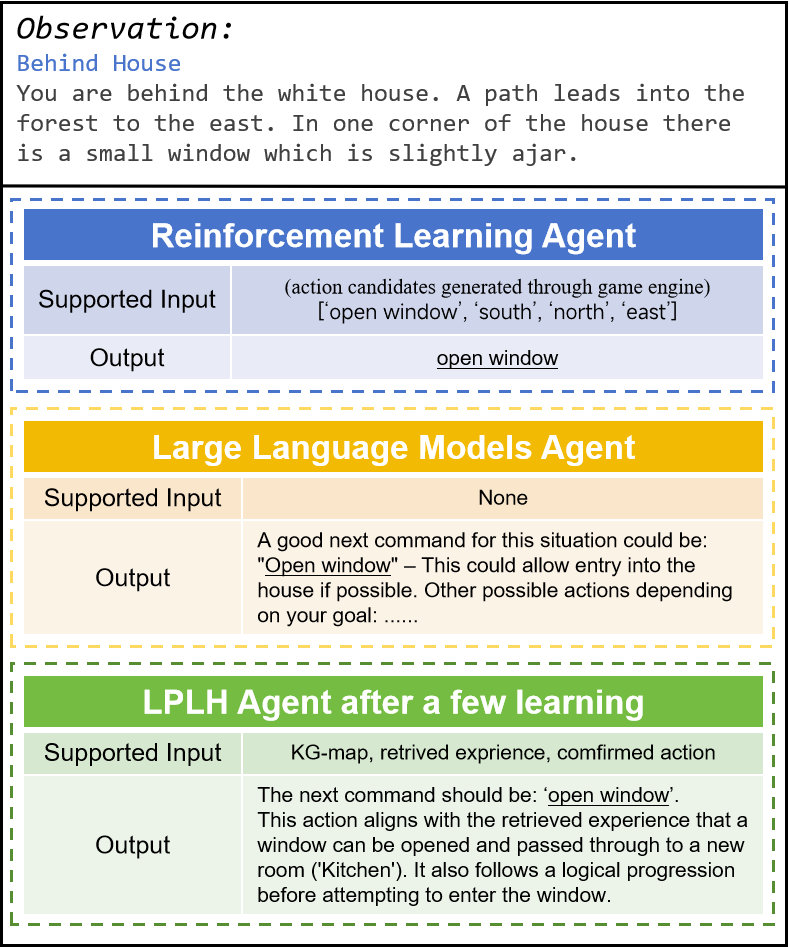}
  \caption{Example of RL approach, the basic LLM approach, and our LPLH approach}
  \label{fig:example}
\end{figure}

Deep Recurrent Neural Network (DRNN) \cite{he2016deep} sparked a growing interest in Reinforcement Learning (RL) settings where states and actions are expressed in natural language. Consequently, IF games have become a core testbed for integrating RL and natural language understanding
\cite{guo2020interactive, Hausknecht_Ammanabrolu_Côté_Yuan_2020}. Early RL agents rely heavily on action filters and simplistic policies \cite{yao2020keep, guo2020interactive, ammanabrolu2020avoid}. Although subsequent RL approaches include more sophisticated techniques \cite{ammanabrolu2020avoid, yao-etal-2021-reading, peng2022inherently}, their score-centric objectives still constrain nuanced narrative reasoning. Recent works have begun leveraging large language models (LLMs) \cite{tsai2023can, ma2024agentboard}, but no system LLMs work on playing IF games yet.

Human intuition is central in how players navigate IF games, enabling adaptive engagement with complex, text-based environments. \citet{bartle1996hearts} characterizes diverse player motivations that shape exploratory and strategic behavior, while \citet{koster2013theory} frames “fun” as the pleasure of pattern recognition—a fundamentally intuitive process of mapping novel challenges to learned solutions. Similarly, \citet{tekinbas2003rules} emphasize how well-structured rules and freedom of interaction support creative problem-solving, highlighting the importance of intuitive, context-sensitive reasoning in game design and analysis.

While prior work has often used IF games to benchmark RL capabilities, we instead focus on leveraging LLMs' capacity for human-like, context-aware reasoning to actively play these games. LLMs have shown promise in multi-step inference, narrative comprehension, and decision-making across text-based domains \cite{huang-etal-2024-prompting, zhang-long-2025-mld, xu2022perceiving, singh2021pre, shi2023self}. IF games, with their rich narratives and open-ended interactions, demand precisely the kinds of situational reasoning and adaptivity that human players naturally exhibit—traits often underemphasized in score-centric RL approaches \cite{tsai2023can}.

To bridge this gap, we introduce the \textbf{L}earning to \textbf{P}lay \textbf{L}ike \textbf{H}umans (LPLH) framework: a novel, training-free LLMs-based system designed to mirror human gameplay strategies in IF environments. Rather than treating human-like as a vague aspiration, LPLH operationalizes it through three targeted modules inspired by common human behaviors during gameplay. \textbf{Dynamic map-building} module incrementally constructs an internal spatial model of the game world, akin to how human players sketch or internalize maps to avoid disorientation. \textbf{Action-space learning} captures and remembers verified verbs and manipulable objects, paralleling how humans accumulate an evolving vocabulary of game-relevant actions. \textbf{Experience reflection} module synthesizes prior successes and failures into reusable summaries, enabling experience-driven decision-making. Through this modular design, LPLH uses contextual game information and reflective memory to make informed decisions, tightly coupling human-inspired reasoning with interpretable, functional components.

By integrating these key modules, the LPLH framework entirely forgoes reliance on external knowledge to pre-train agents. Instead, it fosters a self-innovative learning process that closely mirrors how human players approach new games: constructing detailed maps, incrementally exploring valid actions, and reflecting on experience to inform future decisions. We apply this approach to various IF games from Jercho dataset \cite{guo2020interactive}. The main contributions are as follows: 1) We propose the LPLH framework, designed to simulate human players' behaviors, which, to our knowledge, is the first system to work to leverage LLMs for playing IF games. 2) We demonstrate the robustness and efficiency of LPLH framework by playing games on different baselines. 3) Empirical results confirm that modeling human-like decision-making processes enhances LLM performance on IF game tasks, indicating the effectiveness of our human-player simulation strategy.


\begin{figure*}[htbp]
  \includegraphics[width=\linewidth]{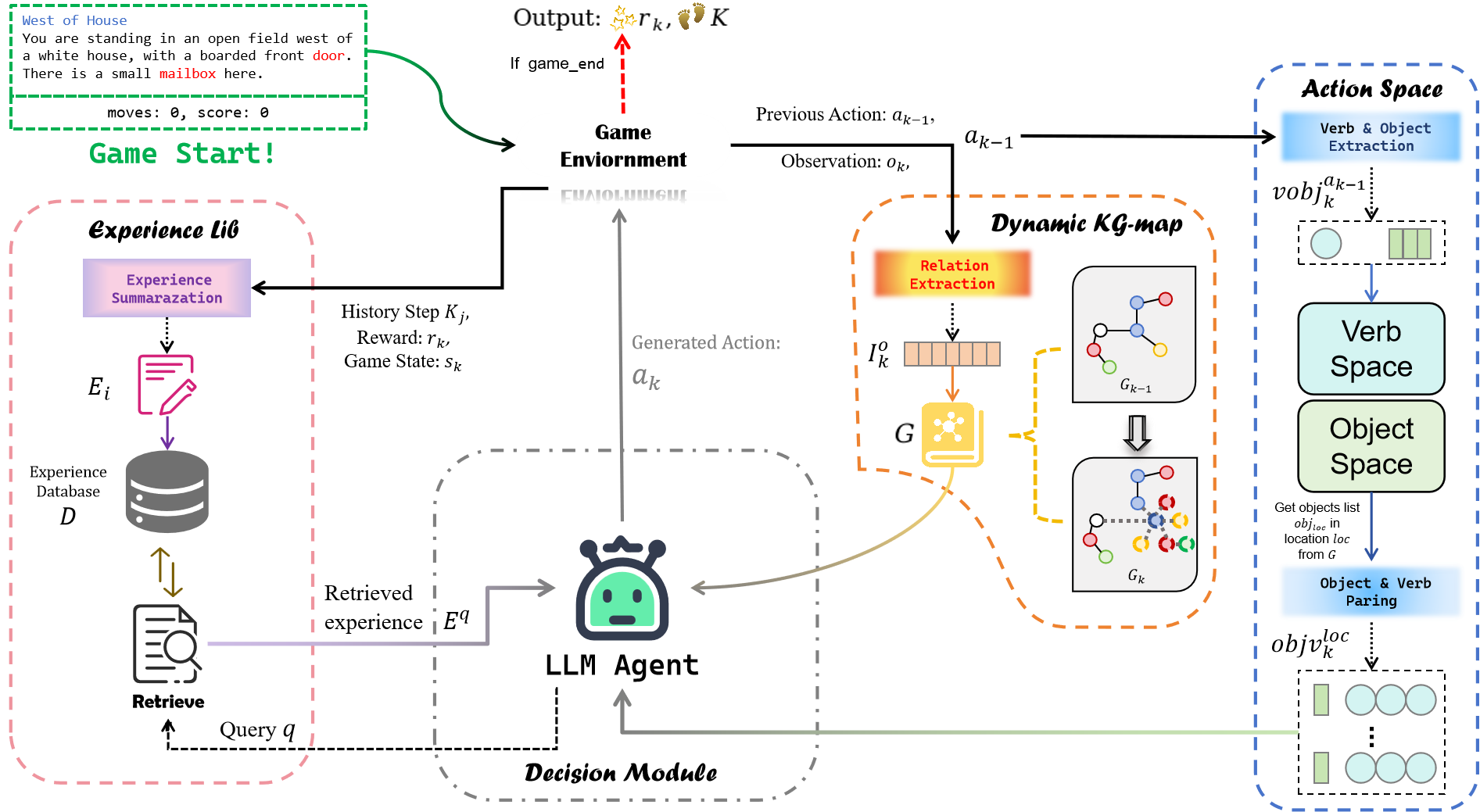}
  \caption {LPLH Framework. The \textbf{Dynamic KG-map} incrementally constructs a knowledge graph from observed items. The \textbf{Action Space} separates valid actions into verb-object pairs for efficient generation. The \textbf{Experience Lib} captures and summarizes key steps as reusable experiences to guide future decisions.} 
  \label{fig: LPLH framework}
\end{figure*}
\section{Related Work}
IF games provide a structured environment that drives research on language-based agents. These text-based simulations \cite{Hausknecht_Ammanabrolu_Côté_Yuan_2020} integrate challenges. Many studies adapt RL methods to handle their vast, partially observable state-action spaces. KG-A2C \cite{ammanabrolu2020graph} builds a knowledge graph (KG) to represent game states and constrain action spaces, addressing the complexity of natural language actions. \citealp{guo2020interactive} reformulates gameplay as a multi-passage reading comprehension task, using context-query attention and structured prediction to enhance action generation and mitigate partial observability.

Another strand of research targets the exploration challenges unique to IF games. Q*BERT and MC!Q*BERT \cite{ammanabrolu2020avoid} employ knowledge-graph-based intrinsic motivation and strategic exploration to overcome bottlenecks in sparse-reward environments. Complementing these efforts, \citealp{tuyls2022multistage} dissects the explore-vs.-exploit dilemma by decomposing each episode into distinct exploitation and exploration phases, achieving notable improvements in normalized game scores across multiple environments.

In parallel to the RL-centric approaches, recent works have explored integrating pre-trained language models to enhance agents' semantic understanding and action generation capabilities. The CALM \cite{yao2020keep} is trained on human gameplay data to produce a compact set of action candidates. CALM significantly improves in-game scores, even on unseen games, when combined with an RL agent for action re-ranking. Similarly, \citealp{singh2022pre} leverages transformer-based models to inject rich semantic priors.

In contrast to approaches that integrate RL, a notable study \cite{tsai2023can} investigates the performance of LLMs on text games without additional RL components. It evaluated ChatGPT \cite{chatGPT} and GPT-4 \cite{openai2024gpt4technicalreport} on Zork1—the only work to focus on leveraging LLMs in this context. Their findings reveal that, although ChatGPT performs competitively with existing systems, it exhibits significant limitations: it fails to construct a coherent model and struggles to incorporate pre-existing world knowledge. These shortcomings highlight critical open questions at the intersection of IF game agents and LLMs, suggesting that further research is needed to fully realize the potential of LLMs-only approaches in interactive fiction environments. Thus, our LPLH framework attempts to fill this gap by simulating human playing behaviors. 

\section{LPLH Framework}
\label{section 3}

Our framework draws direct inspiration from core principles in cognitive science, the interdisciplinary study of how humans perceive, remember, reason, and act~\cite{goswami2008principles}. These principles are particularly relevant in IF games, which demand narrative comprehension, exploration under uncertainty, and intuitive decision-making~\cite{dawson2024evidence}. Rather than treating human-like behavior as a superficial output pattern, LPLH operationalizes it through design choices grounded in cognitive theory.

Specifically, the LPLH architecture simulates spatial reasoning and memory formation via a dynamic knowledge graph that evolves with the agent's exploration—analogous to how human players construct internal or external maps. It mirrors incremental learning and intuitive action formation by continually expanding a repository of validated commands, reflecting how players retain and recombine useful verbs and objects. LPLH also encodes reflective learning through its experience module, enabling the agent to summarize lessons from both success and failure, in line with experiential learning models. Finally, its decision-making process integrates current context, structured memory, and retrieved past experiences—closely paralleling dual-process theories of human reasoning, where both working memory and episodic recall inform adaptive behavior.

In this section, we describe the overall LPLH framework and its core modules in detail. Figure~\ref{fig: LPLH framework} illustrates the system architecture.

\subsection{Problem Define}
The interaction between an autonomous agent and a text-based game environment can be formulated as a Partially Observable Markov Decision Process (POMDP) \cite{spaan2012partially}, represented as a tuple \( (S, T, A, O, R) \) as follows: The agent issues text commands \( a \in A \), selecting from a space of natural language actions; It receives text-based observations \( o \in O \) describing the environment state in a limited scope; The environment provides scalar rewards \( r = R(s, a) \), often sparse, to guide learning; The underlying game state \( s \in S \) encodes KG-map \( G \) but is partially observable through textual feedback; the transition function \( s' = T(s, a) \) updates the game state based on the agent’s action, following the game’s internal logic. 

Unlike the existing RL approaches \cite{he2016deep}, which learns a value function by selecting from a predefined set of actions to maximize game rewards, our LPLH framework more closely mimics human decision-making by integrating multiple sources of information when generating the next command. Specifically, LPLH introduces a structured method for semantic understanding and decision-making in text-based games. At each time step \(k\), LPLH receives the current game observation \(o_k\) and the previous action \(a_{k-1}\) and updates the knowledge graph \(G\) through the dynamic knowledge-graphs map module. Suppose the last action is valid (i.e., the game state changes from \(s\) to \(s'\)); the action space module stores that action by splitting it into its constituent verb and objects. Meanwhile, by evaluating the reward \(r_k\) at each step, LPLH summarizes the current game state \(s_k\) in conjunction with historical information \(K_j\), thus producing a helpful experience \(E_i\), where \(j\) denotes the history length and \(i\) indicates experience index.

For the generation part, our LLMs-based agent \(LLM\) will put the map \(G\), confirmed action list for each object in the current state \(objv\), and retrieved experience \(E\) to predict the best suitable action.

Following sections are details of LPLH framework integrating \textbf{Dynamic knowledge-graphs (KG) map}, \textbf{Action Space}, and \textbf{Experience Lib} enables adaptive and robust learning in interactive fiction environments. This can potentially enhance agent in complex, language-driven tasks.

\subsection{Dynamic knowledge-graphs map}
Creating a KG to store information is widely used for long-term memory solutions in LLM research \cite{tsai2023can, zhu2024llms}. In the IF game task, the KG-map serves as a continually updated map of in-game entities and their relations, thereby guiding the RL agent’s action selection \cite{ammanabrolu2020graph}. However, while previous approaches often treat the KG-map as a static structure or update it only when new object relations are discovered, our dynamic KG-map module continuously modifies the graph after each change. 

Concretely, as the agent interacts with the environment, the textual observations are parsed to capture newly discovered objects, places, or entities and any relationship changes (e.g., “the key is now inside the box”). These updates ensure that the KG-map aligns with the evolving state of the game world. LLMs-based agents can more accurately retrieve relevant information when constructing their following action by maintaining a synchronized, real-time representation of the current environment. This dynamic process resolves inconsistencies (such as outdated item locations) and enriches the agent’s contextual awareness, allowing for more robust decision-making in text-based games.

We employ a \textit{verb \& object extraction}, a fine-tuned model \(fm_{re}\), to identify relational triples (location and objects) from the observation \(o_k\) as relation extraction. Formally, we define:
\begin{equation}
    I^o_k = fm_{re}(a_{k-1}, o_k)
    \label{equ: relation extraction}
\end{equation}
where \(a_{k-1}\) is the action taken at the previous step, and \(I^o_k\) denotes the set of extracted relations.

Subsequently, the module integrates these newly extracted relations \(I^o_k\) along with the preceding action \(a_{k-1}\) to update the knowledge graph \(G\):
\begin{equation}
    G_k =  kg(G_{k-1}, a_{k-1}) \oplus  kg(G_{k-1}, I^o_k) 
    \label{equ: kg update}
\end{equation}
where \(kg(\cdot)\) is a dynamic function that incrementally updates the knowledge graph based on the provided information, and the operator \(\oplus\) combines the updated states from both the action and the newly extracted relations.

\subsection{Action Space Learning}

LPHP framework learns all valid actions within a dedicated action space to emulate human player behavior. This space is decomposed into two phases: verb and object. The valid actions learned in this manner are retained as executable commands. Specifically, after observing the last action \(a_{k-1}\), a \textit{verb \& object extraction} model \(fm_{vo}\) determines whether it remains valid based on the updated observation \(o_k\). If \(a_{k-1}\) is valid, the model decomposes it into a set of verbs and objects:

\begin{equation}
\begin{aligned}
    vobj^{a_{k-1}}_k &= fm_{vo}(a_{k-1}) \quad \text{iff } a_{k-1} \text{ is valid}, \\
    AS &= AS \,\cup\, vobj^{a_{k-1}}_k
\end{aligned}
\end{equation}
where \(AS \in \mathbb{R}^{\{n,m\}}\) denotes the recognized action space, and \(n\) and \(m\) represent the maximum numbers of verbs and objects, respectively, in the game environment. The term \(vobj^{a_{k-1}}_k\) contains exactly one verb (e.g., \textit{“put * in *”}) followed by a list of corresponding objects.

In the subsequent reasoning phase, the framework then employs an \textit{object \& verb pairing} procedure to integrate objects in the current location (\(obj^{loc}\)) with candidate actions:

\begin{equation}
    objv^{loc}_k = \mathrm{pairing}\bigl(obj^{loc}, AS\bigr)
\end{equation}
where the function \(\mathrm{pairing}(\cdot)\) searches the recognized action space to find all actions compatible with current location’s objects. The \(objv^{loc}_k\) is a list of viable action–object pairs for decision-making.

\subsection{Experience Lib}

A core aspect of human gameplay is the ability to reflect on past experiences and reuse knowledge to improve future performance~\cite{5626eec0-6959-3874-807a-ef4f1d09918a}. To simulate this reflective behavior, we introduce an Experience Library that captures and organizes game trajectories into reusable insights. This component is powered by an LLMs-based experience summarization module, which is invoked upon scoring events—either gains or losses—to extract structured, task-relevant information from recent game history.

Given a scoring event, the model identifies the relevant location, traces the key puzzle-solving actions, notes the triggering action and reward change, and distills generalizable insights that may inform future decisions. For instance, a successful trajectory may yield patterns such as “retrieve key from Attic to unlock door in Room1,” while failure cases produce reflective strategies like “examine lantern before entering Basement.” These concise summaries emulate how human players internalize lessons from both success and failure.

Formally, we denote the summarization module as \(LLM_{es}\), which generates a structured experience summary \(E_i\) based on a fixed-length interaction history \(K_j\), associated reward change \(r_k\), and the current game state \(s_k\):
\[
    E_i = LLM_{es}(K_j, r_k, s_k),
\]
where \(i\) indexes the experience and \(j\) indicates the length of historical context. The model is prompted in a one-shot fashion with a structured template to ensure consistency.

Each summarized experience \(E_i\) is stored in a vector database \(D\) for efficient retrieval. During gameplay, we adopt a retrieval-augmented generation (RAG) approach~\cite{lewis2020retrieval} to incorporate these past experiences into decision-making. Given a query \(q\), the system retrieves a relevant subset \(E^q \subset D\) to condition the LLM’s next action. This mechanism grounds the agent’s decisions in prior experience, enhancing factual accuracy, coherence, and adaptability. By strategically reusing previously successful or cautionary trajectories, the agent emulates human-like learning and dynamically refines its policy over time.

\subsection{Zore-shot Decision-making}
To generate the next command \(a_k\) at step \(k\), LPLH use an LLMs-based agent, denoted by \(LLM_a\), in a zero-shot prompting setup. Specifically, the agent first constructs a query \(q\) based on the current game context and uses it to retrieve a relevant experience set \(E^q\). It then aggregates the current observation \(o_k\), the retrieved experiences \(E^q\), the structured KG-map representation \(G\), and the confirmed action set \(objv^{loc}_k\). This combined context is provided as input to the LLMs-based agent, which produces the next command \(a_k\):
\begin{equation}
    a_k = LLM_a\bigl([G,\, objv^{loc}_k,\, E^q],\, o_k\bigr)
\end{equation}
By integrating observations and relevant knowledge, LPLH framework enables the agent to issue commands in a flexible, zero-shot \cite{kojima2022large} way without task-specific fine-tuning.

The proposed LPLH framework integrates zero-shot prompting, retrieval of relevant experiences, action parings, and KG-map structures to guide LLMs-based agents in command generation. This approach offers a robust and adaptable solution that can seamlessly accommodate diverse contexts by eliminating the need for fine-tuning. In doing so, it is the first system methodologies attempt for IF game environments through LLMs-driven techniques,  starting the way for more human-behavior-aligned agent interactions.

\section{Experiment}
This section will briefly introduce our chosen \textbf{Dataset}, \textbf{Baselines}, and \textbf{Experiment setup}.

\subsection{Dataset}
We evaluate our method on a collection of IF games made available through Jericho, an open-source Python-based environment \cite{guo2020interactive}. The games in Jericho cover diverse genres (e.g., dungeon crawl, mystery, horror) and include both classic Infocom titles (like \textit{Zork1}) and community-developed works (such as \textit{Afflicted}). Most IF games employ a point-based scoring system, which serves as the primary reward signal for learning agents. While Jericho natively supports scoring detection for a curated set of games, it also offers the flexibility to run unsupported games without these features. While all games run under \textit{'verbose'} model, which always gives the maximum observation of room. 

\subsection{Baselines}
We choose some previous RL models to compare with different LLMs approaches. These models represent advancements in integrating structured knowledge representations and natural language processing techniques to improve agent performance and interpretability in complex environments. The chosen RL models are \textbf{DRRN} \cite{he2016deep}, \textbf{KG-A2C} \cite{ammanabrolu2020graph} and 
\textbf{DBERT-DRRNL} \cite{singh2022pre}.





Also, several LLM models will be the baseline and foundation for the LPLH framework: \textbf{Qwen2.5-7B-Instruct} \cite{qwen2.5, qwen2}, \textbf{Qwen2.5-14B-Instruct} \cite{qwen2.5, qwen2}, \textbf{GPT-4o-mini} \cite{OpenAI_GPT4oMini_2025} and \textbf{GPT-o3-mini} \cite{OpenAI_O3Mini_2025}.





Noticeably, RL approaches select the possible action candidates supported by the game engine. Meanwhile, all LLM approaches will generate action, the processes of which are more complicated but closer to the human player's ways.



\subsection{Experiment setup}
We assess the proposed LPLH framework on 9 games of varying difficulty levels \cite{guo2020interactive}. Each LLMs-based agent runs for 10 epochs (250 steps per epoch). At every step, only the observation from the Jericho game engine is provided to the agent\footnote{For all RL agents, they receive completed observation and inventory at each step. However, LLMs-based agent needs to decide when to call the command 'look' or 'i' to get such information by themselves.}. We record both the average and maximum score after all epochs for each game. As a comparison, we also implement an LHLP framework under the same experimental conditions but designate only the final three epochs as “learning outcomes,” with all preceding epochs serving as intermediate training phases. Please see Appendix \ref{sec:app:parameters} for hyper-parameters and prompt details.

For the fine-tuned model \(fm\) used in our LHLP framework (Section \ref{section 3}), we adopt a smaller model, \textbf{Qwen2.5-1.5B-Instruction} \cite{qwen2.5}, to address three specific tasks: (1) validating actions, (2) extracting relations from observations, and (3) decomposing actions into verbs and objects. We begin by collecting game actions and observations from the LLM baseline. Next, we employ carefully crafted prompts for GPT-4, obtaining hundreds of annotated samples, which we subsequently refine manually to create the training datasets for each task. Details of the fine-tuned model \(fm\) are provided in Appendix \ref{sec:app:FM reustl}. While the experience summarizing model \(LLM_{es}\) is using \textbf{GPT-o3-mini}.

\begin{table*}[htpb]
    \centering
    \normalsize
    \renewcommand{\arraystretch}{1.4} 
    \resizebox{\linewidth}{!}{
    \begin{tabular}{lcccccccccccc}
        \hline\hline
        \textbf{} & {\textbf{DRRN}*} & {\textbf{KG-A2G}*} & {\textbf{D-D}*} 
        & \multicolumn{2}{c}{\textbf{Qwen-7B}} & \multicolumn{2}{c}{\textbf{Qwen-14B}} & \multicolumn{2}{c}{\textbf{GPT-4o-mini}} & \multicolumn{2}{c}{\textbf{GPT-o3-mini}} & \textbf{Max} \\
        \text{} & \text{} & \text{} & \text{} & \textit{base} & \textit{LPLH} & \textit{base} & \textit{LPLH}  & \textit{base} & \textit{LPLH}  & \textit{base} & \textit{LPLH} & \text{} \\
        \midrule\hline
        
        omniquest &\textcolor{blue}{5} / - &3 / - &4.9 / \underline{5} & 1 / \underline{5} & \textcolor{blue}{5} / \underline{5} & 1.5 / \underline{5} &\textcolor{blue}{5} / \underline{5} &2 / \underline{5} &\textcolor{blue}{5} / \underline{5} &4 / \underline{5} &\textcolor{blue}{5} / \underline{5} &\textbf{50} \\
        
        detective &197.8 / - &\textcolor{blue}{207.9} / - &- / - & 10 / 10 & 68 / \underline{100} & 36 / 70 &72 / 90 &22 / 30 &30 / 60 &20 / 20 &50 / 60 &\textbf{360} \\
        
        zork1  &32.6 / - &34 / - &\textcolor{blue}{44.7} / \underline{55} &0 /0  &9 / 15 & 9 / 35 &39.7 / 45 &6 / 10 &10 / 15 &30 / 35 &33.8 / 45 &\textbf{350} \\
        
        zork3  &0.7 / - &0.1 / - &0.2 / \underline{4} & 0 / 0 & 0.6 / 1 & 2.0 / 3 &2.6 / 3 &1.8 / 3 &2.8 / 3 &\textcolor{blue}{3} / 3 &\textcolor{blue}{3} / 3 &\textbf{7} \\
        
        ludicorp  &13.8 / - &\textcolor{blue}{17.8} / - &12.5 / \underline{18} & 1 / 1 & 1/ 1  & 10.5 / 12 &11.7 / 13 &1 / 1 &2.6 / 3 &4.4 / 7 &8 / 11 &\textbf{150} \\
        
        balances  &\textcolor{blue}{10} / - &\textcolor{blue}{10} / - &- / - & 0 / 0 & 5 / 5 & 8.75 / \underline{10} &\textcolor{blue}{10} / \underline{10} &5 / 5 &5 / 5 &8.3 / \underline{10} &\textcolor{blue}{10} / \underline{10} &\textbf{51} \\
        
        spellbrkr  &37.8 / - &21.3 / - &38.2 / 40 & 0 / 0 & 25 / 25 & 25 / 40 &41.7 / \underline{60} &18 / 40 &38.3 / 50 & 31.3/ 50 &\textcolor{blue}{47.5} / \underline{60} &\textbf{600} \\
       
       dragon    &-3.5 / - &\textcolor{blue}{0} / - &- / - &-3.5 / -1 &-1.3 / 0 &-0.8 / 0 &-0.67 / 0 &-0.8 / -0.2 &\textcolor{blue}{0} / 0 &-4 / -2 &-0.5 / \underline{1} &\textbf{25} \\
        
        gold    &0 / 0 &- / - &- / - &0 / 0 &0 / 0 &2.4 / \underline{3} &\textcolor{blue}{3} / \underline{3} &2.5 / \underline{3} &\textcolor{blue}{3} / \underline{3} &1 / \underline{3} &\textcolor{blue}{3} / \underline{3} &\textbf{100} \\
        
        \hline\hline
        
    \end{tabular}
    }
      \caption{
          Game score results running on IF games. The DRRN*, KG-A2G*, and D-D* (DBERT-DRRNL) are RL agent; results are from their papers. \textit{base} in LLMs-based agent generates action directly with some previous history, and \textit{LPLH} is our approach. For the scores '-/-,' the first represents a \textbf{raw} score of the end, and the second represents the \textbf{max} score. In LLMs-based agents, the raw on \textit{base} computes the average score in all runs, while the raw on \textit{LPLH} computes the last three runs as "learning outcomes." Scores with \textcolor{blue}{blue} mean the highest score in raw, and scores with \underline{underline} are the highest score in max. The \textbf{Max} is the game's maximum score. 
        }
    \label{tab:main result}
\end{table*}

\section{Result and Analyzes}
This section assesses how the LPLH framework advances performance in IF games. First, we illustrate learning behaviors through the learning curves (Section \ref{sec:Learningcurves}), compare game scores across various baselines (Section \ref{sec:game scores}), and conduct an ablation study (Section \ref{sec:ab}). These analyses highlight LPLH’s adaptive, human-like language acquisition capacity and robust performance.


\begin{figure}[ht]
  \includegraphics[width=\columnwidth]{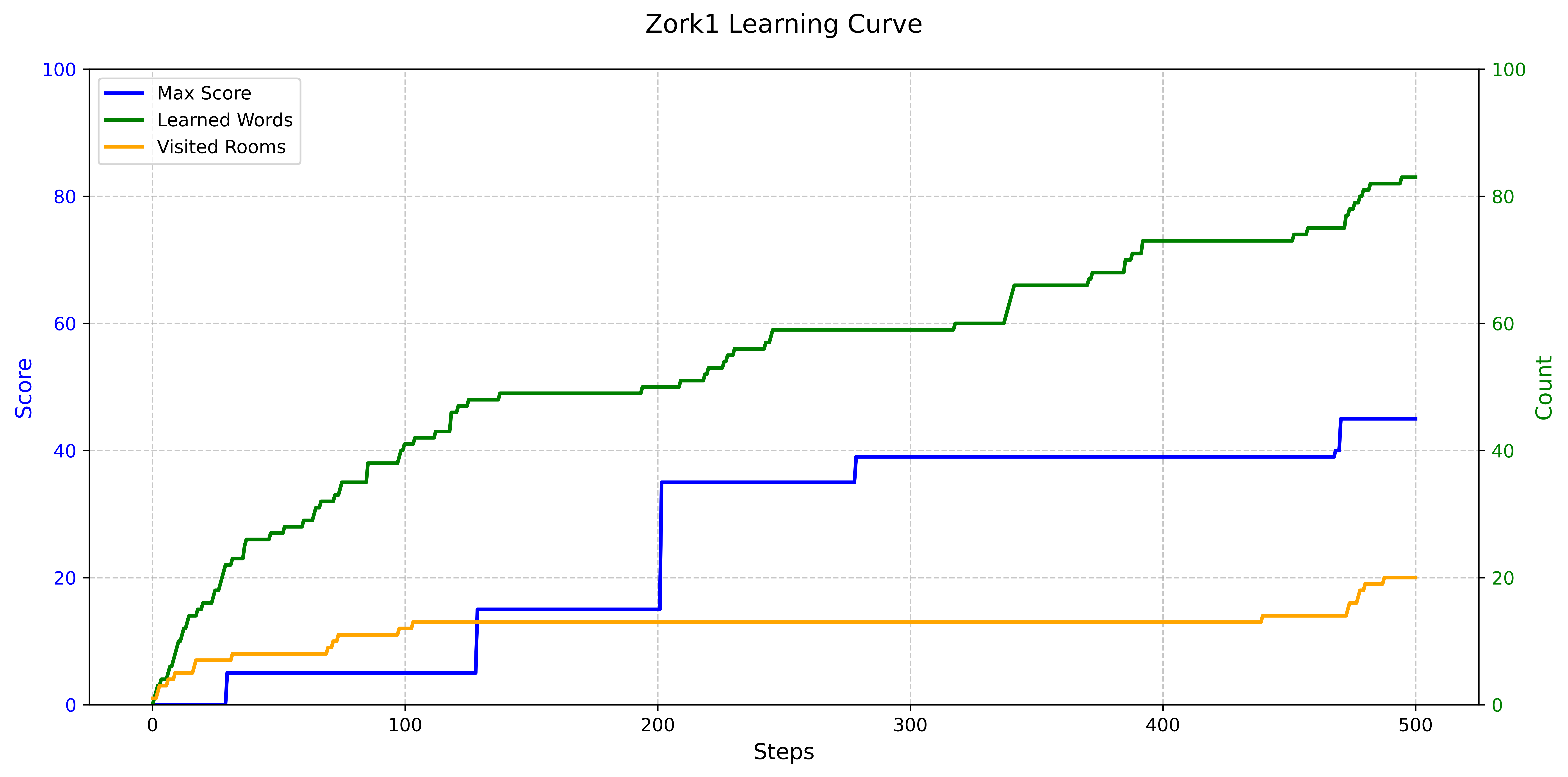}
  \caption{\textit{Zork1} learning curve in scaled steps. For reference, human player's best trajectory gets 350 scores in 412 steps with 48 verbs, 57 objects (total 105 unique words), and 63 rooms.}
  \label{fig:learning_cruve}
\end{figure}

\subsection{Learning Curves}
\label{sec:Learningcurves}
Figure~\ref{fig:learning_cruve} presents LPLH agent’s progression in \textit{Zork1}, with three metrics reflecting key behavioral dimensions: score gains, vocabulary growth, and spatial exploration. While these surface-level patterns resemble trends observed in human players. To strengthen this further, we explicitly ground our analysis in cognitive science and learning theory.

The max score curve (blue) follows a stepwise trajectory, which aligns with the cognitive notion of \textit{insight-driven learning}, where progress occurs in bursts after solving key subgoals~\cite{koster2013theory}. Plateaus reflect periods of trial-and-error, a common phase in strategic human exploration. The learned actions curve (green) shows early rapid acquisition followed by gradual slowing, consistent with language learning theories that posit an initial burst of accessible verbs followed by slower mastery of complex interactions. Similarly, the visited rooms curve (orange) rises steeply during early exploration and later flattens, reflecting an exploration-to-exploitation shift—a behavior well-documented in models of human decision-making under uncertainty~\cite{boyd1983reflective}. This reveals a positive association between accumulated knowledge and task success, echoing findings in educational psychology that emphasize iterative reflection and action refinement as drivers of improved performance~\cite{5626eec0-6959-3874-807a-ef4f1d09918a}. 

Together, these patterns and forthcoming analyses suggest that LPLH exhibits not merely functional success but learning dynamics that parallel human cognitive strategies—reflecting the theoretical foundations upon which the framework is built.

\subsection{Game Scores}
\label{sec:game scores}

Table~\ref{tab:main result} summarizes score performance of various agents on multiple IF games, comparing previous RL approaches with LLMs-based methods. The LPLH framework markedly improves the performance of LLMs-based agents by enabling dynamic adaptation and iterative learning during play.

Across the board, we observe consistent and substantial gains when integrating LPLH into LLMs-based agents. For example, in \textit{Detective}, Qwen-7B (LPLH) achieves a surprising 68/100 compared to the base model’s 10/100, representing a 6.8× raw improvement. Similarly, in \textit{Spellbrkr}, Qwen-14B (LPLH) attains 41.7/60, outperforming both its base version and the RL agent DBERT-DRRN (D-D). These results highlight LPLH’s capacity to leverage learned experiences akin to human players refining their strategies through trial and error.

Notably, LPLH agents can match the maximum scores of strong RL baselines in certain games. For instance, in \textit{Omniquest} and \textit{Balances}, both o3-mini (LPLH) and the best-performing RL agents reach the game's upper score bound, with progress converging at known puzzle bottlenecks\footnote{See section~\ref{sec:erroranalysis} Error Analysis for more details.}~\cite{ammanabrolu2020avoid, tuyls2022multistage}. This indicates that LPLH can achieve comparable task completion without explicit reward shaping or extensive environment-specific tuning.

Our results suggest that LPLH framework enhances agents by fostering deeper contextual understanding and dynamic strategy adjustments beyond static priors. Moreover, LLMs-based agents with LPLH typically require fewer steps than RL methods to achieve comparable or higher scores \cite{guo2020interactive}, all without relying on additional external knowledge. As shown in Figure \ref{fig:example}, LPLH agents exhibit human-like reasoning steps, providing a clear, self-explanatory rationale for their actions. Future work may focus on further optimizing LPLH to reinforce adaptive behavior in more complex IF settings, potentially bridging the gap between fine-grained RL solutions and the flexible, learned knowledge of LLMs-based approaches.

\begin{table}[htbp]
\centering
\small
\renewcommand{\arraystretch}{1.2}
    \begin{tabular}{p{3.5cm} c c c}
    \hline\hline
    \textbf{Model}      & \textbf{raw}  & \textbf{max}  & $\sigma$ \\
    \hline
    $\text{LPLH}_{14B}$          & 39.7              & 45.0              &   4.2       \\
    \hline
    \hspace{0.5em}$\text{LPLH}_{14B}-\textit{CoT}$    & 41.6       & 45.0         &  2.4        \\
    \hspace{0.5em}KG-map only             & 11.0             & 15.0             &  2.0         \\
    \hspace{0.5em}KG-map + exp               & 11.0             & 35.0             &  13.1        \\
    \hspace{0.5em}KG-map + as               & 27.8             & 35.0             &  6.8        \\
    \hspace{0.5em}exp only     & 25.6             & 34.0             &  9.0         \\
    \hspace{0.5em}exp + as       & 32.0              & 40.0              &  4.0         \\
    \hspace{0.5em}as-only   & 26.6              & 35.0              &  6.6         \\
    \hline\hline
    Qwen2.5-14B-Instruct                 & 9.0             & 25.0            & 9.2      \\
    \hspace{0.5em}14B-\textit{select one}   & 14.5      & 30.0    & 11.6 \\

    \hline

    \hline\hline
    \end{tabular}
\caption{Ablation results on '\textit{Zork1}.' Where $\sigma$ is the standard deviation; 'exp' represents the experience summarization; and 'as' represents the action space. }
\label{tab:ablation study}
\end{table}

\subsection{Ablation Study}
\label{sec:ab}
In this section, we provide a more in-depth analysis of LPLH framework and evaluate contribution of each component. As shown in Table \ref{tab:ablation study}, we report both the raw and max scores, as well as their standard deviations (\(\sigma\)) to assess performance stability across different model variants. We take LPLH\textsubscript{14B}\footnote{The backbone model is Qwen2.5-14B-Instrction.} as our backbone model and observe that fine-tuning it with chain-of-thought reasoning (LPLH\textsubscript{14B}-CoT\footnote{We use DeepSeek-R1-Distill-Qwen-14B \cite{deepseekai2025deepseekr1incentivizingreasoningcapability} as the CoT distillation model for this task.}) boosts the raw score from 39.7 to 41.6, while maintaining a relatively small standard deviation, 2.4. However, this improvement comes at a higher computational cost.

Next, we examine the effects of adding a knowledge graph mapping component (KG-map). Although this variant exhibits a slightly lower maximum score, its standard deviation is reduced, suggesting improved stability. The model achieves a higher maximum score when additional experiential data are introduced and exhibits a larger \(\sigma\). Finally, incorporating an action-space mechanism provides further performance gain by reducing wasted actions. Combining this mechanism with experiential data leads to the most substantial overall results, demonstrating the effectiveness of a multi-faceted approach to enhancing LPLH.

We also compare against a Qwen2.5-14B-Instruct baseline (9 raw, 25 max, 9.17 \(\sigma\)) and observe further gains by incorporating a selective mechanism (14B-\textit{select one}\footnote{The selective mechanism is choosing one action from game engines' action candidates same as RL approaches.}). Each module  independently boosts performance, and their synergy yields even stronger results. Future work includes exploring how transitioning from generating to selecting actions may further enhance reasoning.


\section{Discussion experiences' significance in LPLH framework}
This section discusses two examples in \textit{Zork1} to show how experiences are essential for the LPLH framework to simulate humans. Both are how LPLH framework plays like humans with experience reflection to achieve higher scores and avoid failure, where we will use 'player' to represent the game character and 'agent' for LPLH framework.

\subsection{Learning from Failure}
In the case of how the LPLH framework learns from failure, the player meets the dark environment for the first time. The agent would not figure out what was going on and then try to take any new action, but a death followed anyway. After the player's death, the agent will automatically start summarizing this experience. During the summarization, agent will also focus on any partially missing events and suggestions for future reference. After the first attempt, the agent calms that \textit{`You may need lights to avoid death in the dark...`}. However, the agent suggests finding lights since the player never finds light resources. After several attempts, the player finally finds a 'lantern' in the 'living room' and takes it. When a player goes to dark again, the experience will let the agent know that the light needs to be turned on first.

\subsection{Learning from Success}  
Learning from success is more straightforward to analyze. Once the player solves a puzzle, the pertinent steps are captured and organized to guide future decision-making. Consequently, when the player encounters a similar situation, the agent retrieves these validated experiences and follows the proven trajectory of actions. Once the player obtains an “egg” at “Up a Tree,” they earn five points during the initial exploration. In subsequent rounds, the agent consistently instructs the player to collect this “egg” at the start of the game. “egg” and “Up a Tree” are automatically stored in the KG-map, facilitating quick retrieval for future scenarios.

In LPLH framework, each interaction, success or failure, provides essential feedback that informs subsequent decisions \cite{schaul2015prioritized, browne2012survey}. By systematically archiving and reflecting on these experiences, the agent refines its understanding of the environment, thereby improving strategic behavior and adaptability over time \cite{bion2023learning, boyd1983reflective}.


\section{Error Analysis: Known Puzzle Bottlenecks}
\label{sec:erroranalysis}
Despite LPLH’s improvements over baseline methods, RL- and LLMs-based approaches struggle with specific puzzle bottlenecks. This limitation stems mainly from the difficulty of discovering domain-specific or unconventional commands rarely seen in training data or standard exploration trajectories, which also has been pointed out in \citet{tuyls2022multistage, ammanabrolu2020avoid}.

In IF games, key progress often hinges on executing highly specific actions that are not inferable from the immediate context. For instance, in \textit{Zork1}, players must type \texttt{echo} in the “Loud Room” to earn five points—a solution that eludes most automated agents, whether they sample from a predefined action space (e.g., Jericho-based RL systems) or generate free-form text via LLMs. These methods typically lack the inductive bias or experiential grounding to propose such obscure commands.

Even with structural guidance from a dynamic knowledge graph, the agent may overlook necessary commands. In one representative case from the “Living Room,” progress depends on typing \texttt{move rug} to reveal a hidden passage. The agent frequently fails here—substituting implausible variants like \texttt{hit rug} or ignoring the rug altogether—though it occasionally succeeds with \texttt{move rug} or \texttt{pull rug}.

These examples illustrate a fundamental challenge in IF games: successful progression often depends on domain-specific intuition, contextual knowledge, or cultural cues that are not easily recoverable from surface-level patterns. As a result, both RL- and LLMs-based agents still fall short of reliably solving such puzzles, limiting their ultimate performance despite improvements in exploration and memory modules.
Addressing this gap remains a crucial direction for future work, underscoring the need to align agents with intuition and imagination that guide human players.

\section{Conclusion}

In this work, we introduced \textbf{L}earning to \textbf{P}lay \textbf{L}ike \textbf{H}umans framework, \textbf{LPLH}, a novel framework designed to guide LLMs to play IF games by simulating human player behaviors. To our knowledge, this is the first systemized approach leveraging LLMs to tackle well-known text-based IF games. This cognitively inspired approach for LLMs-based IF game agents integrates dynamic map building, action space learning, and experience-driven memory. LPLH framework tries to balance narrative comprehension, exploration, and puzzle-solving by simulating human play processes without relying on external pre-training. Although our approach still falls short of specialized RL agents in certain games and cannot match human-level scores, it yields more interpretable, human-like behaviors and enables more context-aware decision-making in interactive fiction game domains. Furthermore, according to the performance of LLMs in IF games, we believe that IF games are a considerable challenge for LLMs in many aspects. 

\section*{Limitation}
To our knowledge, LPLH framework is the first attempt to enable LLMs to play IF games by simulating human players as a system works. Although we incorporate multiple modules for long-term memory, effectively managing and navigating that memory remains challenging. Currently, our approach uses a simple experience summarization that is only triggered when the agent loses or gains points. In contrast, human players naturally integrate relevant information into their memory at any point during gameplay, suggesting that a more dynamic summarization strategy could yield better results.

Furthermore, the framework relies on a JSON-structured KG-map as input. The consistency and clarity of this representation can influence the model’s reasoning, indicating that further investigation is needed to determine the optimal representation method for LLMs-based agents in IF tasks. We also attempted to evaluate the LPLH framework across different models; however, we could not perform extensive tests on a wide range of LLMs due to resource constraints. Future work should include more comprehensive experimentation and exploring adaptive memory-management techniques to address these limitations. Also, we only test a few IF games to show our framework performance, which may not be fully adopted for all IF game types. During the game, the learning process is still affected by many factors that could dramatically lead to score increases or decreases, which we have not found. 

Also, the current framework is only applied to IF games. We'll try to fit LPLH in wider text-based environments such as TextWorld~\cite{cote2019textworld}.

\section*{Acknowledgments}
This work is supported by the Alan Turning Institute/DSO grant: Improving multimodality misinformation detection with affective analysis.

\bibliography{custom}

\appendix

\section{Fine-tuned model \(fm\)}
\label{sec:app:FM reustl}
Before we use the fine-tuned model \(fm\) in LPLH framework. We collect the data through three different games (not in our test game): '\textit{Dragon},' '\textit{Karn}', and '\textit{Night}.' 

In both games, we run with random pick action provided by the game engine and generate the action through LLMs. Then, we pair the action and sequenced observation as the basic training dataset. For this basic training dataset, we manually delete those repeat parts. Following that, we create different prompt templates for three task: (1) validating actions (Prompt in Table \ref{tab: Prompt template: Action validation}), (2) extracting relations from observations(Prompt in Table \ref{tab: Prompt template: Relation Extraction}), and (3) decomposing actions into verbs and objects (Prompt in Table \ref{tab: Prompt template: Splitting Action}). After getting the results generated from GPT-4o, we manually selected the correct parts and then passed them to the train. We use LoRA \cite{hu2021lora} to train the model on LLaMA-Factory \cite{zheng2024llamafactory}. The training details are shown in Table \ref{tab:hyper-parameters of fine-tune}. 

We evaluate the model performance by running '\textit{Zork1}' with 'walk-thought'\footnote{the human player's best trajectory}. For the task of validating actions, the accuracy is 90\%. For relations extraction, the error rate is like 15\%. And for splitting the actions, the accuracy is 98\%.

\section{Baseline Details}

\textbf{DRRN} \cite{he2016deep} models the relevance between the state and possible actions to navigate large action spaces effectively.

\textbf{KG-A2C} \cite{ammanabrolu2020graph} integrates dynamically constructed KG into the Advantage Actor-Critic framework to constrain the action space and improve decision-making.


\textbf{DBERT-DRRNL} \cite{singh2022pre} enhances the traditional DRRN architecture by incorporating DistilBERT \cite{sanh2020distilbertdistilledversionbert}, a pre-trained language model, to provide richer text representations, thereby improving the agent's performance.

\textbf{Qwen2.5-7B-Instruct} \cite{qwen2.5, qwen2}: A 7B open-source model from Alibaba, designed for general-purpose natural language understanding and generation, optimized for efficiency and broad-domain applicability.

\textbf{Qwen2.5-14B-Instruct} \cite{qwen2.5, qwen2}: A larger 14B version of Qwen, offering improved reasoning, generation quality, and contextual understanding.

\textbf{GPT-4o-mini} \cite{OpenAI_GPT4oMini_2025}: A lightweight version of GPT-4o, optimized for efficiency while maintaining strong reasoning capabilities, making it suitable for scalable applications.

\textbf{GPT-o3-mini} \cite{OpenAI_O3Mini_2025}: A compact version of OpenAI’s third-generation model, designed for high-speed inference with reasonable performance in various NLP tasks, especially in constrained computational environments.

\begin{table}[htbp]
    \centering
    \footnotesize
    \begin{tabularx}{\linewidth}{X|X}
    \textbf{Parameter name} & \textbf{Value} \\
    \hline
        \verb|lora_rank| \rule{0pt}{2ex}    & $16$   \\ \hline
        \verb|lora_alpha| \rule{0pt}{2ex}  & $32$  \\ \hline
        \verb|lora_dropout|\rule{0pt}{2ex} & $0.1$ \\ \hline
        \verb|lora_target|\rule{0pt}{2ex}  &   \texttt{all}    \\ \hline
        \verb|learing rate|\rule{0pt}{2ex} & $2e-5$ \\ \hline
        \verb|epoches|\rule{0pt}{2ex}      & $3$   \\
    \hline
    \end{tabularx}
    \caption{hyper-parameters of fine-tuning}
    \label{tab:hyper-parameters of fine-tune}
\end{table}

\section{Hyper-parameters and Prompts}
\label{sec:app:parameters}
\subsection{Hyper-parameters}
All experiences were running at 2 \(\) RTX4090 GPUs with torch type of \textit{bf16}. ALL none-trained LLMs (\(LLM_{es}\) and \(LLM_a\)) run with the same temperature of 0.6. The fine-tuned model (\(fm\) uses a temperature of 0.1.

\subsection{Prompt templates}
Here, we show some essential prompt templates, which all models followed CoT reasoning. Table \ref{tab: Prompt template: Baseline} shows how the baseline model generates the next command. Table \ref{tab: Prompt template: experience sum} shows a prompt template of how to summarize the experience. The prompt template of action generation of the LPLH framework is showing Table \ref{tab: Prompt template: LPLH action generation}. In our study, no any game name or specific game commands appear in all prompts. According to research done by \citep{tsai2023can}, the Chat-GPT knows IF game \textit{Zork1}. When Chat-GPT knows the specific games, it will do well. 

\begin{table*}[htbp]
    \centering
    \footnotesize
    \begin{tabularx}{\linewidth}{X}
    \hline
    \textbf{Instruction:}\rule{0pt}{2ex} \\
You are evaluating the outcome of a text-based game action based on the game’s observation (feedback message) after the player’s previous action. Your task is to determine if the action was successful or not.\\

<START OF INSTRUCTIONS>\\
- You will be given an observation text that follows the player’s attempted action.\\
- If the observation indicates that the action was carried out successfully (e.g., it provides new information, describes the environment, or gives a positive confirmation), respond with:\\
  <ais> True </ais>\\
- If the observation indicates that the action could not be performed (e.g., includes phrases like "You can't..." or "You cannot..."), respond with:\\
  <ais> False </ais>\\

Note:\\
- An unsuccessful action usually explicitly states that the player cannot do something, or that the action fails.\\
<END OF INSTRUCTIONS>\\
    \hline
    \end{tabularx}
    \caption{Prompt template: Action Validation.}
    \label{tab: Prompt template: Action validation}
\end{table*}

\begin{table*}[htbp]
    \centering
    \footnotesize
    \begin{tabularx}{\linewidth}{X}
    \hline
    \textbf{Instruction:}\rule{0pt}{2ex} \\
<START OF INSTRUCTIONS>\\
You're going to extract triples in the format <subject, relation, object> from an input Observation along with previous actions you did, originating from a text-based game. Focus solely on where the character ('You') is located, what objects are in that location, and their immediate properties. The maximum length for any object name in the triples is three words, where length of location name has no limit.\\

Rules:\\
1. If the observation doesn't describe an environment or information is insufficient (e.g., "Opened", "Taken"), output |start| none |end| and skip other points.\\

2. Always use 'in' as the relation to represent the character's location. Convert any spatial descriptions (e.g., 'are facing', 'are standing', 'are behind') to the 'in' relation. If the input begins with a Room name (starts with a capital letter and does not end with a period), use it as the location.\\
   Example: \\
   Input: "Stairwell (First Floor) You're in the north stairwell."\\
   Triple: <You, in, Stairwell (First Floor)>\\

3. If the observation doesn't include a precise location, do not provide any <You, in, *> triple.\\

4. Use 'have' as the relation to represent interactive objects present in the location. Focus only on the objects themselves as the 'obj' in the triple. Ignore decorative details unless they indicate an interactive object. Limit object names to a maximum of three words.\\
   Example:\\
   Input: "There is a small mailbox here."\\
   Triple: <[Location], have, mailbox>\\

5. Do not include additional details or properties of objects. Only extract the objects themselves, ensuring object names are no longer than three words. But if a object have a relation to another object, such as 'in' and 'on', then extract that relation.\\
   Example:\\
     Input: "A buzzing water fountain has been moved."\\
     Triple: "<[Location], have, water fountain>"\\
     Input: "A sock is on th table."\\
     Triple: "<[Location], have, sock>, <[Location], have, table>, <sock, on, table>"\\

6. If the input specifies a requirement or action needed to continue, use <location/object, need/require, something to action>.\\
   Example:\\
   Input: "Forest. You would need a machete to go further west."\\
   Triple: <Forest, need, machete to go west>\\

7. For objects or locations mentioned with a direction (e.g., 'to the north', 'up to', 'down'), use <current location, direction, [new location]/to [direction]>.\\
   Example:\\
   Input: "Hall. To the southwest is the entrance to the Computer Site, and you can go east here as well as go up with a stair."\\
   Triples: <Hall, southwest, Computer Site>, <Hall, east, to east>, <Hall, up, to up>\\

Note: Pay more attention to objects and directions than to objects' states or other decorative details.\\

Now, extract the relationships for the input step by step and merge all the results into a single output enclosed within |start| * |end|, where * represents the list of extracted triples.\\

<END OF INSTRUCTIONS>\\
    \hline
    \end{tabularx}
    \caption{Prompt template: Relation Extraction.}
    \label{tab: Prompt template: Relation Extraction}
\end{table*}

\begin{table*}[htbp]
    \centering
    \footnotesize
    \begin{tabularx}{\linewidth}{X}
    \hline
    \textbf{Instruction:}\rule{0pt}{2ex} \\
<START OF INSTRUCTIONS>\\
You wil receive a previous input(step) from a text-based IF game, and please split the input into two parts, action and objs, as "<verb; [objs]>". Please follow these instructions to complete the task step by step.\\

Use the following rules:\\
1. If the action is a simple directional command (e.g., "north" or "n"), the object list should be empty.\\
For example:\\
    Input: "west" \\
    Response: "<act> <west; []> </act>" \\
    
2. If the action is "take all" or another "all" command (e.g., "take all"), treat "take all" as the verb and leave the object list empty.\\
For example:\\
    Input: "drop all" \\
    Response: "<act> <drop all; []> </act>" \\
    
3. If there are objects following the Verb (e.g., "eat", "take") or Verb phrase (e.g., "drop down", "go around"), list them. If prepositions (e.g., "on", "at", "with") are present, include them in the verb phrase using "\&" as a placeholder, and list each noun object individually.\\

Final Output:\\
Use <act> <verb; [objs]> </act> format for final output where:
"verb" represents the action phrase with placeholders "\&" for objects. objs is a list of object nouns.\\

<END OF INSTRUCTIONS>\\
    \hline
    \end{tabularx}
    \caption{Prompt template: Splitting Action.}
    \label{tab: Prompt template: Splitting Action}
\end{table*}

\begin{table*}[htbp]
    \centering
    \footnotesize
    \begin{tabularx}{\linewidth}{X}
    \hline
    \textbf{Instruction:}\rule{0pt}{2ex} \\
    
You are playing the classic text-based interactive fiction game. Your goal is to explore, solve puzzles, collect treasures, and reach the winning end state. Throughout the game, you will:\\

1. Receive a history of the game’s the action you performed, the new observation representing what you see or experience after your action. The action will help you understand what led to the new observation.\\
2. Have access only to the last 10 turns of conversation as your history. You must rely on these 10 turns and your internal reasoning to keep track of your location, inventory, puzzles, and progress.\\
3. Receive current new observation based on the last action and the current game states as input.\\
4. Produce all responses formatted between "|start|" and "|end|".\\
\\
**Your Task:**\\
- At each turn, carefully read the provided new observation and the action you performed.\\
- Use your internal chain-of-thought to determine the best possible action to advance in the game, considering your inventory, location, known puzzles, and the limited historical context you have.\\
- Once you have reasoned through your options, produce exactly ONE game command.\\
- Always Format your command as this at the end of your response:\\
**Final Command:**\\
  |start|
  [your chosen command]
  |end|

**Guidelines:**\\
- Avoid random or nonsensical actions. Each move should serve a purpose or help solve a puzzle.\\
- Try to use player (human) logic to guide your decision.\\
- You can Use 'look' command to examine the current location. And 'inventory' command to examine your inventory.\\
- Maintain continuity by leveraging the last 10 turns of conversation and the last action you performed as a guide. Use your internal chain-of-thought to continue.\\
- Always think first, then act. The chain-of-thought is mandatory before producing the final command.\\
\\
Remember, you are playing a text-based game. Follow these instructions diligently, use your chain-of-thought to reason about your actions, and only format your final chosen command between "|start|" and "|end|".\\

    \hline
    \end{tabularx}
    \caption{Prompt template: Baseline action generation.}
    \label{tab: Prompt template: Baseline}
\end{table*}

\begin{table*}[htbp]
    \centering
    \footnotesize
    \begin{tabularx}{\linewidth}{X}
    \hline
    \textbf{Instruction:}\rule{0pt}{2ex} \\

<START OF INSTRUCTIONS>

You are a game engine summarizer. Your task is to read the current log of the game state and produce a concise, cohesive summary of the player's progress so far (This happens every time the player gets a score or loses a score). Do NOT reveal any hidden or undiscovered information. Focus only on details the player already knows or has directly experienced.\\
A list of "Step" will be provided. Each step includes:\\
- An observation (what the player sees),\\
- Info about moves and current score,\\
- The action taken just before the observation.\\

**Summary Structure:**\\
1. "location": where the player is (or what area is described) when the score changes. If the player has died, give the location name before death.\\
	*1.1* - One Location name Only.\\
	*1.2* - Description of situation.\\
2. "puzzle\_status": what puzzles or obstacles have been solved to earn/lose the points.\\
	*2.1* -  ONLY related steps to solve the puzzles directly. Any requirement for solving the puzzles, such as 'player need to <step>open door<step> at Room1 to enter <loc>Room2<loc>.\\
	*2.2* - Description of the puzzle.\\
3. "scoring": how the player earned/lost points for the last step. Any action leads to earning/losing points.\\
	*3.1* - Step done to earn/lose points.\\
	*3.2* - How many points are changed?\\
4. "important\_experience": The experience can be used for the future. Only the most notable and valuable clues or items the player learned about for the global game experience or any warning must be recorded through all previous logs. Only Focus on confirmed information.\\
	*Earn Points* - ONLY when player earn points, then we only need to know what leads to earn points and ingore other unchecked information.\\
	- For example: 'player noticed there is a rabbit on the table (unchecked)' is not experienced. 'Room1. player open a locked door by a key (The key got from the roof)' is the experience as' player need to go to <loc>roof<loc> for a <obj>key<obj> to open the locked door in <loc>Room1<loc>.\\
	*Lose Points* - ONLY when the player loses points (died usually or lost in the game, where 'lost' here means the player earned no points for a long time ), you also need to give suggestions for the future.\\
	- For example: 'player died in Room2. (Player saw a rabbit on table in Room1, but player did nothing with the rabbit)' you can now give the suggestion for next time that try to check <obj>rabbit<obj> before going <loc>Room2<loc>.\\

**Remember**:\\
- If no related puzzles (solve puzzles to show new location or new environment observations or earn points) are encountered, the whole 'puzzle\_status' needs to be "No puzzles encountered yet."\\
- Please focus on how the player scored points with related puzzles and situations that occurred.\\
- Do not reveal hidden or undiscovered info.\\
- Keep it concise and factual based on the logs.\\
- When giving "important\_experience", please reflect like an expert player (Always think about why this happened) as the payer's 'trace game experience'.\\
- If player has not died, the '*Lose Points*' in 'important\_experience' should be 'none'. If player has died, the '*Earn Points*' in 'important\_experience' should be 'none'.\\
- In your reasoning, if you find more than one earning or losing points, please ONLY summarize the last one based on previous steps.\\

**Final Output Format:**\\
- In the final output for any 'loc name', please use <loc> loc name <loc> to mark it, as well as 'step did before' (which steps solved the puzzles) by marking in <step> step did <step>, as well as 'interacted obj' (which player did valid action to obj) by marking in <obj> interacted obj <obj>; where the 'interacted obj' in step doesn't need this marking. And give a structured output based on points.\\
- At the end of the response, please outline TAGs (no more than 4) between <tag> * </tag> that are used for retrieval. put main location in <room> * </room> as one of the tag.\\
- After TAGs, please also give the difficulty for current puzzles in between <dif> * </dif>. You can combine the history steps with your expert player's experience to define the difficulty.
- Please think about it first. Then, give your final completed player experience summary between '|start|' and '|end|'.\\

<END OF INSTRUCTIONS>\\

    \hline
    \end{tabularx}
    \caption{Prompt template: experience summarization.}
    \label{tab: Prompt template: experience sum}
\end{table*}

\begin{table*}[htbp]
    \centering
    \footnotesize
    \begin{tabularx}{\linewidth}{X}
    \hline
    \textbf{Instruction:}\rule{0pt}{2ex} \\
    
<START OF INSTRUCTIONS>\\
**Instructions for Generating a Next Command in Text-Based Interactive Fiction**

---\\

**Objective**  
Craft a single, context-aware **next command** with it's motivation that propels the game forward, based on the current map, recent actions, and history of attempts. This command should represent one immediate player action.

---\\

**Principles for Exploration, Puzzle-Solving, and Earning Points**

1. **Analyze the Current Game State**  \\
   - **Room \& Map Details**: Assess where you are, noting any exits, known layout, and significant objects.  \\
   - **Recent Attempts**: Reflect on the previous actions, the motivation of taking that action and observation after this attempt.  \\
   - **Inventory Check**: Identify items on hand (keys, tools, etc.) that might solve current puzzles or overcome obstacles.  \\
   - **Objects \& Interactions**: Focus on confirmed items or directions. If uncertain leads might advance the game, consider them cautiously.  \\
   - **Action Selection**: Only choose to interact with an object (or perform an action) if you’re confident it will move the story forward.

2. **Use Retrieved Experiences and Past Attempts** \\ 
   - **Relevance**: Apply past successes or observed clues that align with the current room or situation.  \\
   - **Avoid Repetition**: Do not repeat failing commands indefinitely. If a command fails, adjust strategy.  \\
   - **Focus on Gains**: Prioritize moves likely to unlock new paths, uncover essential items, or yield valuable information.

3. **Formulate a Single Effective Command**  \\
   - **One Action**: Provide exactly one executable command.  \\
   - **Purpose**: Briefly ensure it’s the most logical next step, considering both context and success likelihood.\\
   - **Move command**: The full directions are ['north', 'south', 'east', 'west', 'southeast', 'southwest', 'northeast', 'northwest', 'up', 'down'] 

4. **Output Format**  \\
   - Present the final command and a short motivation in the following format without extra commentary:\\
     ```\\
     You internal reasoning steps Here.\\

     |start|\\
     <com>[command]</com>\\
     <rea>[short motivation for the decision-making reason]</rea>\\
     |end|\\
     ```

---\\

**Adaptation and Fallback Rules**

1. **Priority Usage**  \\
   - **Highest Priority**: Items in `temp\_have`.  \\
   - **Next**: Options in `may\_direction` or `may\_have`.  \\
   - **Then**: Verified directions (`direction`) or items (`have`).\\

2. **Conflict Resolution**  \\
   - Disregard prior attempts known to fail at this location or context.  \\
   - Validate uncertain (`may\_`) directions or items before fully committing to them.\\
   - After verify all the exits in one room then you can fully trust the map.\\

3. **Fallback Strategies**  \\
   - If uncertain, explore unvisited areas or re-examine ('look') the current room.  \\
   - Look for overlooked clues or alternative ways forward.\\

4. **Exploratory Commands**  \\
   - If tools are available, think of how to use them on obstacles.  \\
   - In case an exploration fails, attempt a different angle—return to a previous room, look around again, or try another approach.\\
   - **Explore the world**: It's better to try all directions in each room to identify the exit and update the game map. For `may\_direction`, consider testing that path (e.g., “north”).\\
  
---  \\
**Remember**: You are navigating a text-based world. Combine current observations with past knowledge to decide the best single move. \\

<END OF INSTRUCTIONS>\\
    \hline
    \end{tabularx}
    \caption{Prompt template: LPLH action generation.}
    \label{tab: Prompt template: LPLH action generation}
\end{table*}


        
        
        

\end{document}